\documentclass{article}

\usepackage{PRIMEarxiv}
\usepackage[utf8]{inputenc}
\usepackage[T1]{fontenc} 
\usepackage{hyperref}  
\usepackage{url}    
\usepackage{booktabs}    
\usepackage{tabularx}
\usepackage{amsfonts}    
\usepackage{amsmath}     
\usepackage{nicefrac}   
\usepackage{microtype} 
\usepackage{lipsum}
\usepackage{fancyhdr}     
\usepackage{graphicx}      
\usepackage{float}
\graphicspath{{media/}}    
\usepackage{subcaption}
\usepackage{comment}
\usepackage{lineno}
\usepackage{xcolor}

\usepackage[most]{tcolorbox}

\newtcolorbox{promptbox}[2][]{
  breakable,
  colback=gray!5!white,
  colframe=gray!75!black,
  fonttitle=\bfseries,
  title={#2},
  before upper=\fontsize{10}{12}\selectfont,
  #1
}

\title{ACORN: AI CO-Scientist Reporting Norms}

\author{\parbox{0.94\textwidth}{\centering
{\large
Yara Kyrychenko$^{a,b}$ \& Jon Roozenbeek$^{a,c}$}
\\[24pt]
{\small
$^{a}$Department of Psychology, University of Cambridge, Cambridge, UK
\\[4pt]
$^{b}$Centre for Human-Inspired Artificial Intelligence, University of Cambridge, Cambridge, UK
\\[4pt]
$^{c}$Department of Communication Science, Vrije Universiteit Amsterdam, Amsterdam, Netherlands
\\[24pt]
Correspondence: \textmd{yk408@cam.ac.uk, jjr51@cam.ac.uk}}
}}
\date{}

\begin{document}

\maketitle

\begin{abstract}
AI agents and coding tools such as Claude Code and ChatGPT Codex can now carry out chains of unsupervised actions and perform coding and analysis tasks at least on par with junior software developers or experienced research assistants. These developments have major implications for how empirical research is conducted, shifting the burden on human researchers from coding to oversight and review. However, successfully integrating agentic AI tools into scientific research requires robust guidelines to safeguard the accuracy, interpretability, ethical compliance, transparency, and reproducibility of agentic AI and human outputs. We present the AI CO-Scientist Reporting Norms, or ACORN (\href{https://acorn-checklist.github.io/}{acorn-checklist.github.io}), a reporting checklist for research that uses AI agents, compliant with all major scientific publishers’ AI usage guidelines. Designed for any empirical or formal scientific discipline, ACORN records which agent(s) were used, what they did, who (between the agents and the human) made each decision, and how to reproduce outputs. The ACORN checklist is drafted by the AI agent, and reworked and approved by the human. We release ACORN as an agent skill that we expect to evolve over time. To this end, we provide a dedicated website for community deliberation on individual items and components and invite all researchers to submit their suggestions. We discuss the implications of agentic AI use in scientific research and hope that ACORN can facilitate the study of AI-assisted science.
\end{abstract}

\vspace{2.5em}

\newpage
\section*{Introduction}
Since the launch of OpenAI’s ChatGPT in 2022, generative AI has grown rapidly in both its adoption and abilities. Researchers across fields have turned to these tools as research aides, for example, for automated content classification in the social sciences \cite{feuerriegel_using_2025}. More recently, AI use in research has gone well beyond classification to autonomous or semi-autonomous research creation \cite{lu_towards_2026}. Tools like Claude Code, ChatGPT Codex, and Gemini Deep Research can substantially speed up the process of moving from a research idea to a completed study–gradually becoming indispensable for literature review, experimental software development and data analysis and visualisation–and may also improve reproducibility by generating, annotating, and interpreting research code \cite{Miao2026ReimaginingAgents}. Increasingly, even the most complex intellectual challenges, such as the Navier-Stokes Millennium Prize Problem and the Riemann Hypothesis, are reportedly falling to the advanced internal models of AI companies \cite{Ward2026AIsolved,Olivares2026}.

However, while these tools are improving researchers' productivity \cite{kusumegi2025scientific}, they also come with new challenges. Part of the burden falls on academic publishers, editors, and peer reviewers, who must ensure that the large quantity of AI-assisted research passes rigorous quality checks. To this end, major publishers have instituted various AI policies to govern and disclose how AI is used in research. For instance, every one of the top five academic publishers by the number of journals, which together had over 10,000 journals and 2,240,000 articles in 2025 \cite{Scilit2026}, has published a detailed AI policy demanding transparent disclosure and outlining prohibited uses of AI. Yet disclosure formats vary from author to author, and AI use often goes unreported despite publisher policies \cite{he_bu_2026_academic}, making it difficult to evaluate the extent of AI involvement and contribution to academic work. 

Moreover, when working with AI agents, researchers need to be mindful not to outsource key decisions of the analytical process to the model, which may not only behave sycophantically (and thus bend towards pleasing the user even if they are making mistakes; \cite{Rathje2025SycophanticOverconfidence}) but also hallucinate or misunderstand key research concepts. Outsourcing important components of scientific research risks offloading not only analyses but also their interpretation to AI, which may lead to reduced understanding on the part of researchers, as well as a “monocultural” approach to asking and answering scientific questions \cite{Traberg2026AIMonoculture}. 

AI agents also exacerbate the risks of research malpractice, as they can conduct many different analyses quickly, giving researchers more degrees of freedom. A lack of transparency in reporting such analyses would undermine research integrity. In one study, 71 research teams were given the same dataset (on the relationship between immigration and public support for public welfare programmes) and were free to choose their own analysis plan \cite{Borjas2026IdeologicalFindings}. This led to wide variation in researchers’ analytical decisions and results, but more importantly, evidence of ideological bias: pro-immigration research teams tended to find positive impacts of immigration on support for social welfare, whereas anti-immigration teams tended to find the opposite. The authors argue this occurred because different teams adopted different model specifications. With the rise of agentic AI, this may be particularly problematic because AI agents make undeclared assumptions about their human users, for example with respect to their perceived ideological stance \cite{Tornberg2026PoliticalAuditor}. These findings suggest that researchers' degrees of freedom may have a substantial impact on what conclusions end up being reported in the scientific literature–a problem that is likely to increase with increased AI assistance.

We summarise some opportunities and challenges of using agentic AI in scientific research in Table \ref{tab:table-1}.

\begin{table}[h]
\caption{Key opportunities and challenges of agentic AI for empirical research.}
\label{tab:table-1}
\centering
\renewcommand{\arraystretch}{1.4} 
\begin{tabularx}{\linewidth}{@{}
>{\hsize=1.01\hsize\raggedright\arraybackslash}X>{\hsize=0.99\hsize\raggedright\arraybackslash}X@{}}
\toprule
\textbf{Opportunities} & \textbf{Challenges} \\ \midrule
Speed of analysis & Lack of time for oversight \\
Scale of analysis; enabling more robustness checks & Lack of knowledge for oversight; risk of undue offloading \\
Standardisation: Fewer researcher degrees of freedom, explicit design choices, and mandatory assumption tests & Inappropriate modelling or statistical decisions for unique datasets; risk of scientific “monoculture”; sycophancy and undeclared assumptions by the agent \\
Reproducible code & The illusion of reproducibility (other areas might not be reproducible, such as study design) \\
Accessibility (API calls do not require compute) & Privacy risks and GDPR violations \\ \bottomrule
\end{tabularx}
\end{table}
\newpage

\section*{ACORN: a reporting checklist for agentic AI in science}
As agentic AI becomes increasingly integrated into researchers' routines, it is essential that such integration be reported transparently to ensure rigour, reproducibility and ethical integrity. To address this, we present ACORN (AI CO-Scientist Reporting Norms), a reporting checklist designed to increase transparency in agentic AI use for science. ACORN aims to encourage responsible and transparent use of agentic AI in research by providing a shared disclosure standard. 

In designing the checklist, we initially drew inspiration from GUIDE-LLM, a recent consensus-based reporting checklist for using large language models in behavioural and social science research \cite{feuerriegel_reporting_2026}, which nevertheless does not cover the use of AI agents for research. We adapted GUIDE-LLM items to the agentic AI context and ensured compliance with the top five academic publishers’ AI policies (Elsevier, Springer Nature, Taylor \& Francis, Wiley, SAGE; top five by the number of journals in 2025 according to Scilit), as well as two additional major publishers (AAAS, which publishes \emph{Science}, and PNAS). To minimise adoption barriers, ACORN is designed to be easily integrated into agentic AI workflows (e.g., as an agent skill). To complete the checklist, researchers can simply prompt their model to create a draft based on their project's conversational history. The agent would produce a draft of the checklist in a machine-readable format, from which it can generate a human-readable Word, LaTeX, or R Markdown file. The machine-readable format also lets one easily combine and analyse completed checklists across research papers, which we hope will enable a rigorous quantitative study of how AI is transforming the research process.

\begin{figure}[t!]
    \centering
    \includegraphics[width=\linewidth, clip]{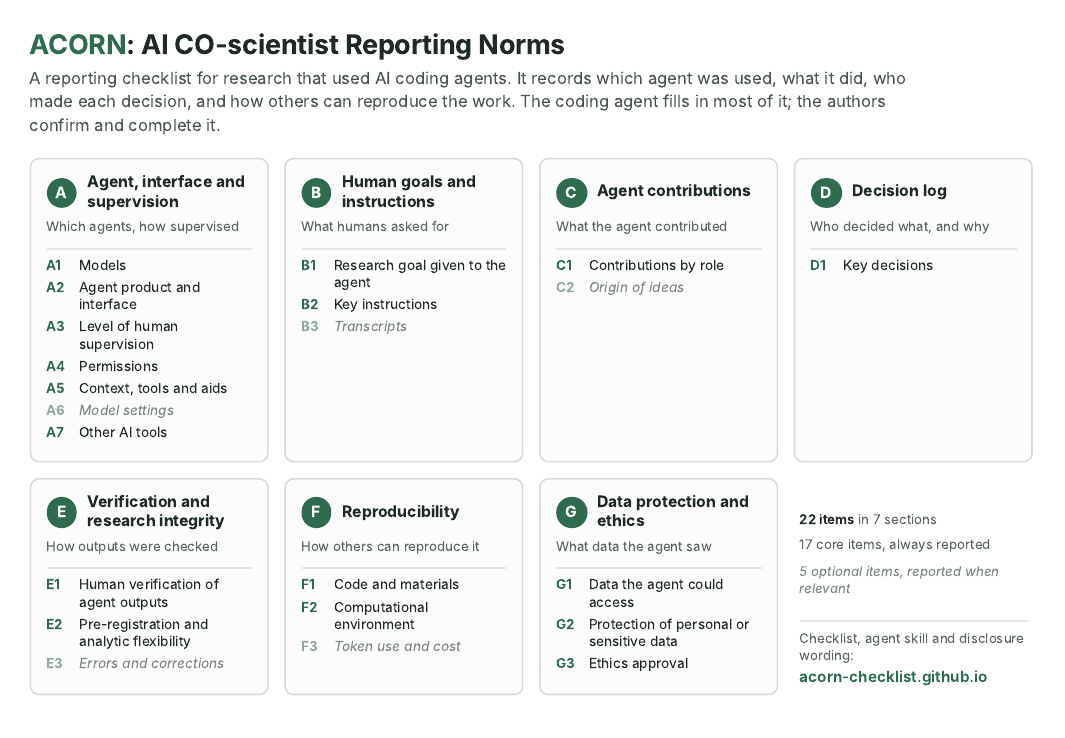}
    \caption{Overview of the ACORN checklist. This figure was created with the help of an AI coding agent (Claude Code with Claude Opus 5.5, by Anthropic), which wrote the script that draws it from the ACORN checklist file. The authors checked that it accurately represents the checklist.}
    \label{fig::ACORN}
\end{figure}

The current version (v0.1.0) of the checklist is organised into seven sections (A-G) with 17 core and 5 optional items (see Figure \ref{fig::ACORN} for an overview). The checklist is available online as a living standard where anyone can contribute a suggestion or a discussion point for any item via GitHub: \href{https://acorn-checklist.github.io/}{acorn-checklist.github.io}. Changes to existing items, such as to item wording, guidance, or answer option edits, and suggestions for items to be added, merged or removed should be accompanied by a brief explanation for why the change is necessary. Contributors who choose to provide an ORCID iD will be credited appropriately if their suggestions are accepted. We encourage all researchers to provide constructive critiques and suggestions for AI reporting norms via the website and will review proposed changes on a rolling basis.

Section A of the checklist reports on the details of the AI agent or agents used, such as model ID and context like skills. Section B discloses any human instructions provided to the agent during the project, optionally pointing to the anonymised conversational transcripts of researchers. Section C outlines agent contribution by research role according to the CRediT contributor taxonomy. Because AI coding tools make many decisions independently and confidently, ACORN specifies where and why the human-in-the-loop agreed or disagreed with its decision-making in Section D. Section E discloses the extent of researcher verification of AI output, as well as any pre-registration and analytical flexibility. In addition, ACORN makes reproducibility easier by encouraging the agent to report the necessary code, materials, and computational environments in Section F. Finally, Section G covers data protection and ethics approval for data processing with an AI agent (if applicable). Importantly, the final responsibility for all aspects of the research and the completed checklist always rests with the human authors. We summarise the above in Figure \ref{fig::ACORN}. 

\section*{Conclusion}
The rapid adoption of agentic AI in scientific research has come with immense opportunities and major discoveries, including solutions to long-standing complex problems in mathematics and physics. However, AI models’ penchant for sycophancy \cite{Rathje2025SycophanticOverconfidence} and propensity to make key decisions without user input \cite{Tornberg2026PoliticalAuditor} also come with substantial risks for the robustness, replicability, and integrity of research outputs. Addressing these challenges requires robust AI reporting norms across scientific disciplines. 

Here, we present ACORN, the AI CO-Scientist Reporting Norms. ACORN is designed to make it as easy as possible for researchers to report how they used agentic AI in their work, which decisions were made by themselves and the model, and how the research can be reproduced. The ACORN checklist can be easily compiled by the agent and verified and approved by the human within the agentic environment; the resulting report can then be included as part of a study appendix or in an online repository. Importantly, ACORN is compliant with AI use guidelines set by seven major scientific publishers. To ensure that ACORN is future-proof, the website (\href{https://acorn-checklist.github.io/}{acorn-checklist.github.io}) includes a community deliberation feature where researchers can provide input about how the tool can be used and improved. Overall, we call for the widespread adoption of transparent agentic AI reporting norms across the scientific community. 

\section*{Acknowledgements}
An AI coding agent (Claude Code, versions 2.1.326, 2.1.331 and 2.1.332, with the Claude Opus 5.5 model, claude-opus-5-5; Anthropic) assisted with developing ACORN: it drafted the checklist items from original author input, as well as guidance and documentation, wrote the software and website, mapped publisher AI policies to ACORN items and made the overview figure of the checklist, all under the authors' direction. The agent came up with the acronym ACORN. The authors read all outputs, one author (J.R.) independently tested the agent skill, and the authors take full responsibility for the content. The authors wrote the manuscript; the agent drafted this disclosure statement, which the authors reviewed and updated. Grammarly was used for language editing. The use of the agent is documented in the completed ACORN checklist (v0.1.0) \cite{kyrychenko2026acorn}, available at \href{https://doi.org/10.5281/zenodo.24560705}{doi.org/10.5281/zenodo.24560705}.

\section*{Author contributions}
Y.K. and J.R. conceptualised the study and wrote the original draft. Y.K. led the development of ACORN.

\section*{Declarations}
The authors declare no competing interests.

\newpage
\bibliographystyle{naturemag}  
\bibliography{references,zotero}  
\end{document}